\documentclass[journal]{IEEEtran}
\usepackage[utf8]{inputenc}

\usepackage{algorithm}
\usepackage{algorithmic}
\usepackage{amsmath}
\usepackage{amssymb}
\usepackage{array}
\usepackage{bm}
\usepackage{booktabs}
\usepackage{cases}
\usepackage[pdftex]{graphicx}
\usepackage{indentfirst}
\usepackage{mathrsfs}
\usepackage{multirow}
\usepackage{url}
\usepackage{xcolor}
\usepackage{makecell}
\usepackage{pifont}
\usepackage{color}  
\usepackage{subfigure}
\usepackage{underscore}
\usepackage{soul}

\usepackage[colorlinks,
            linkcolor=red,
            anchorcolor=black,
            citecolor=green]{hyperref}
\newcommand{\eg}{\textit{e.g.},~}
\newcommand{\ie}{\textit{i.e.},~}

\begin{document}

\title{A Survey for Deep RGBT Tracking}
%
\author{Zhangyong~Tang,
        Tianyang~Xu,
        and Xiao-Jun~Wu 
\thanks{Z. Tang, T. Xu and X.-J. Wu are with the School of Artificial Intelligence and Computer Science, Jiangnan University, Wuxi, P.R. China. (e-mail: \{zhangyong\_tang\_jnu; tianyang\_xu; xiaojun\_wu\_jnu\}@163.com)}

}
%
%
%
%
\maketitle
\sloppy
\begin{abstract}
Visual object tracking with the visible (RGB) and thermal infrared (TIR) electromagnetic waves, shorted in RGBT tracking, recently draws increasing attention in the tracking community. 
Considering the rapid development of deep learning, a survey for the recent deep neural network based RGBT trackers is presented in this paper.
Firstly, we give brief introduction for the RGBT trackers concluded into this category.
Then, a comparison among the existing RGBT trackers on several challenging benchmarks is given statistically.
Specifically, MDNet and Siamese architectures are the two mainstream frameworks in the RGBT community, especially the former.
Trackers based on MDNet achieve higher performance while Siamese-based trackers satisfy the real-time requirement.
In summary, since the large-scale dataset LasHeR is published, the integration of end-to-end framework, \eg Siamese and Transformer, should be further considered to fulfil the real-time as well as more robust performance.
Furthermore, the mathematical meaning should be more considered during designing the network.
This survey can be treated as a look-up-table for researchers who are concerned about RGBT tracking.

\end{abstract}
%
%
%

\section{Introduction}\label{introducion}
\begin{figure}[ht]
	\begin{center}
		\includegraphics[width=1\linewidth]{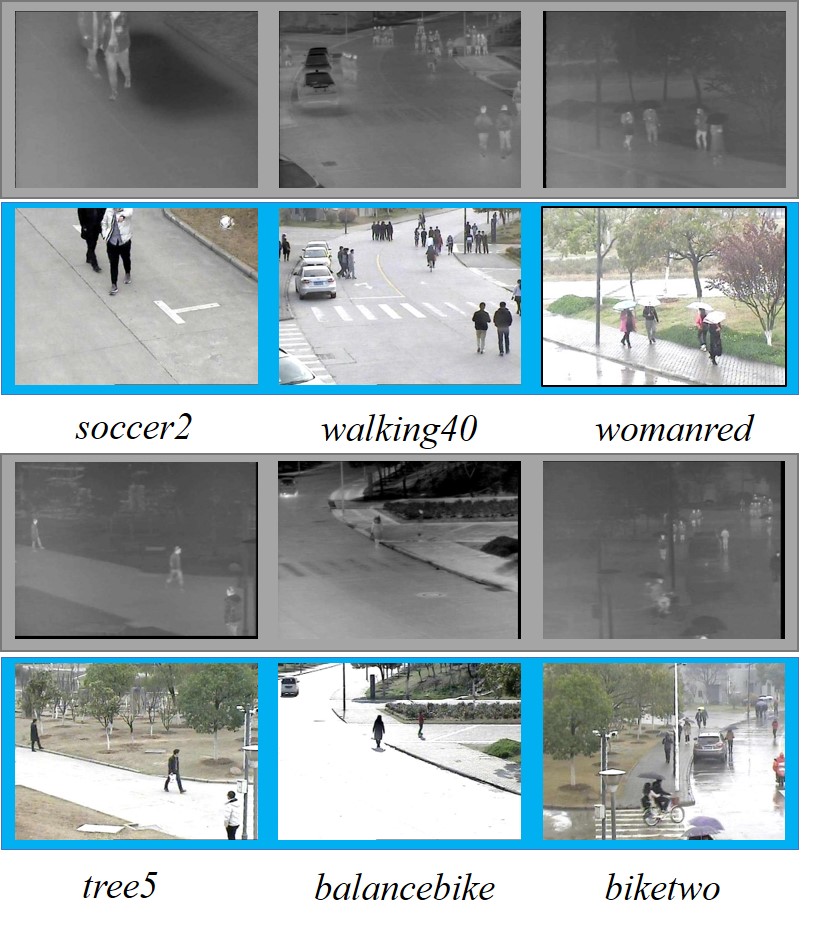} 
	\end{center}
    \caption{An illustration of paired RGBT data. Here the sequences \ie \textit{soccer2, walking40, womanred, tree5, balancebike, biketwo} are selected from RGBT210 \cite{RGBT210}. The region in gray indicates the TIR images while the other represents images in RGB space.}
    \label{fig:rgbt}
\end{figure}

Given the first frame as the prior, visual object tracking aims at predicting a compact bounding box of the object among the whole sequence.
Due to the price friendly characteristic of visible sensor, RGB tracking occupies the dominating status in the tracking community.
However, current researches have noticed its shortcomings in extreme environment conditions, \eg fog and night.
Basically, RGB data is imaged with the electromagnetic waves reflected by the object.
Thus, it is persuasive that all the factors interfering the reflection have direct influence on the imaging procedure.
On the contrary, thermal data is imaged by the emitted thermal radiations from the objects with temperature above absolute zero.
Therefore, it is considered being complementary to the RGB modality.
Recently, thanks to the development of RGB and TIR all-in-one equipments, the multi-modal clues can be constructed simultaneously and applied to the downstream tasks, \eg tracking \cite{zhu2020complementary, ZhuTCSVT, LADCF, GFSDCF}, image fusion \cite{li2020nestfuse, luo2017image, li2017multi, luo2016novel} and segmentation \cite{Li2015CVPR-seg, Li2016TIP-seg}.
In this paper, the main concentration is put on tracking with multi-modal, \ie RGBT tracking.
Fig. \ref{fig:rgbt} shows some pairs of RGBT data.

The rest of this paper is arranged as follows.
We briefly introduce the existing deep RGBT trackers in Section \ref{trackers}.
And then their quantitative performances are compared on several benchmarks in Section \ref{datasetsandresults}.
Then the conclusion is given in the final Section \ref{conclusion}.

\section{RGBT Trackers}\label{trackers}
In this section, we will introduce the mainstream frameworks employed in the RGBT tracking community.

\begin{figure*}[ht]
	\begin{center}
		\includegraphics[width=1\linewidth]{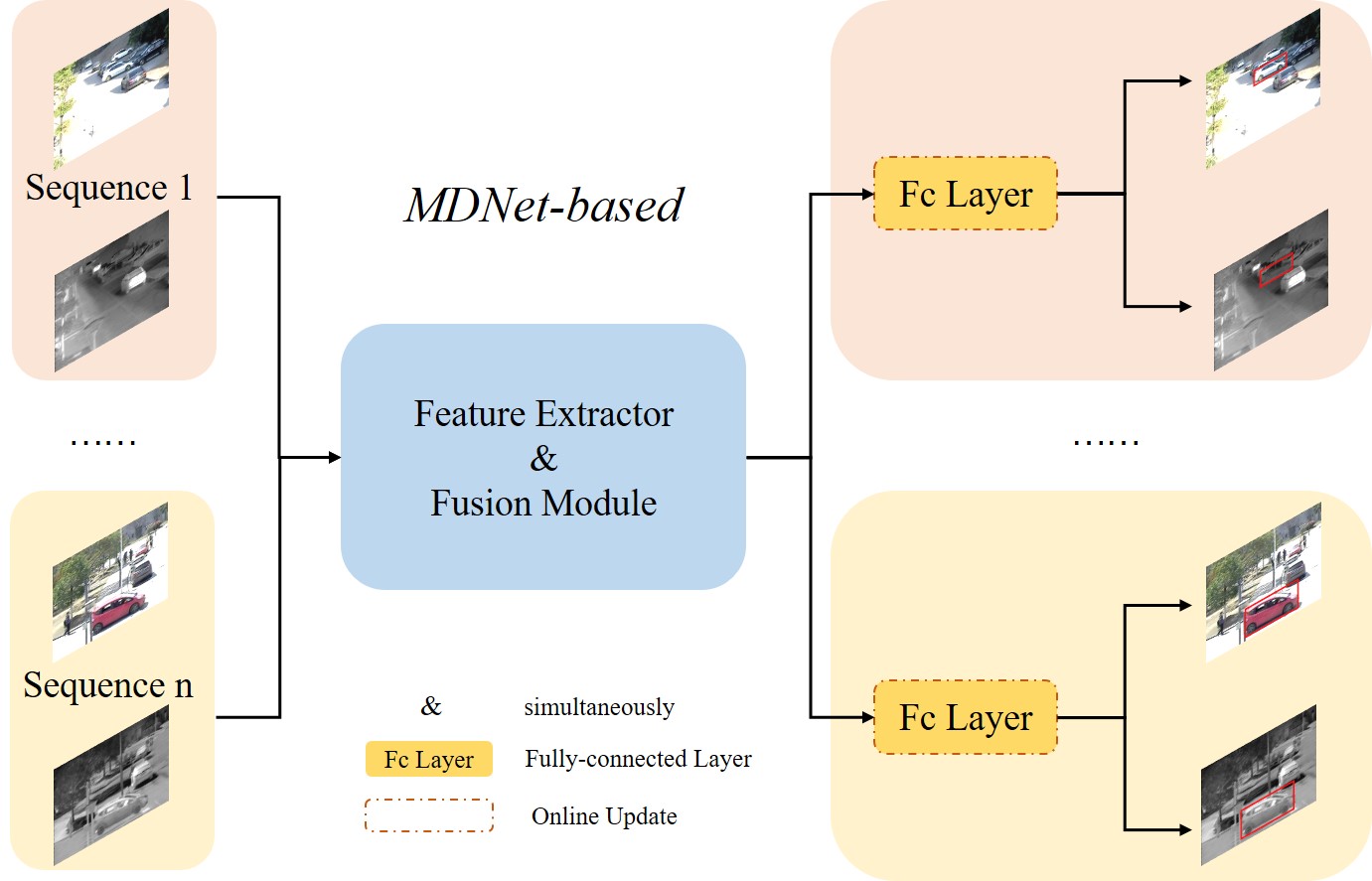} 
	\end{center}
    \caption{Pipeline of MDNet-based RGBT trackers. The processes of feature extraction and multi-modal fusion are completed in the first blue block. And the block rendered in yellow is the last fully-connected (Fc) layer, which is online initialized before the tracking procedure.}
    \label{fig:mdnet-rgbt}
\end{figure*}

\begin{figure*}[ht]
	\begin{center}
		\includegraphics[width=1\linewidth]{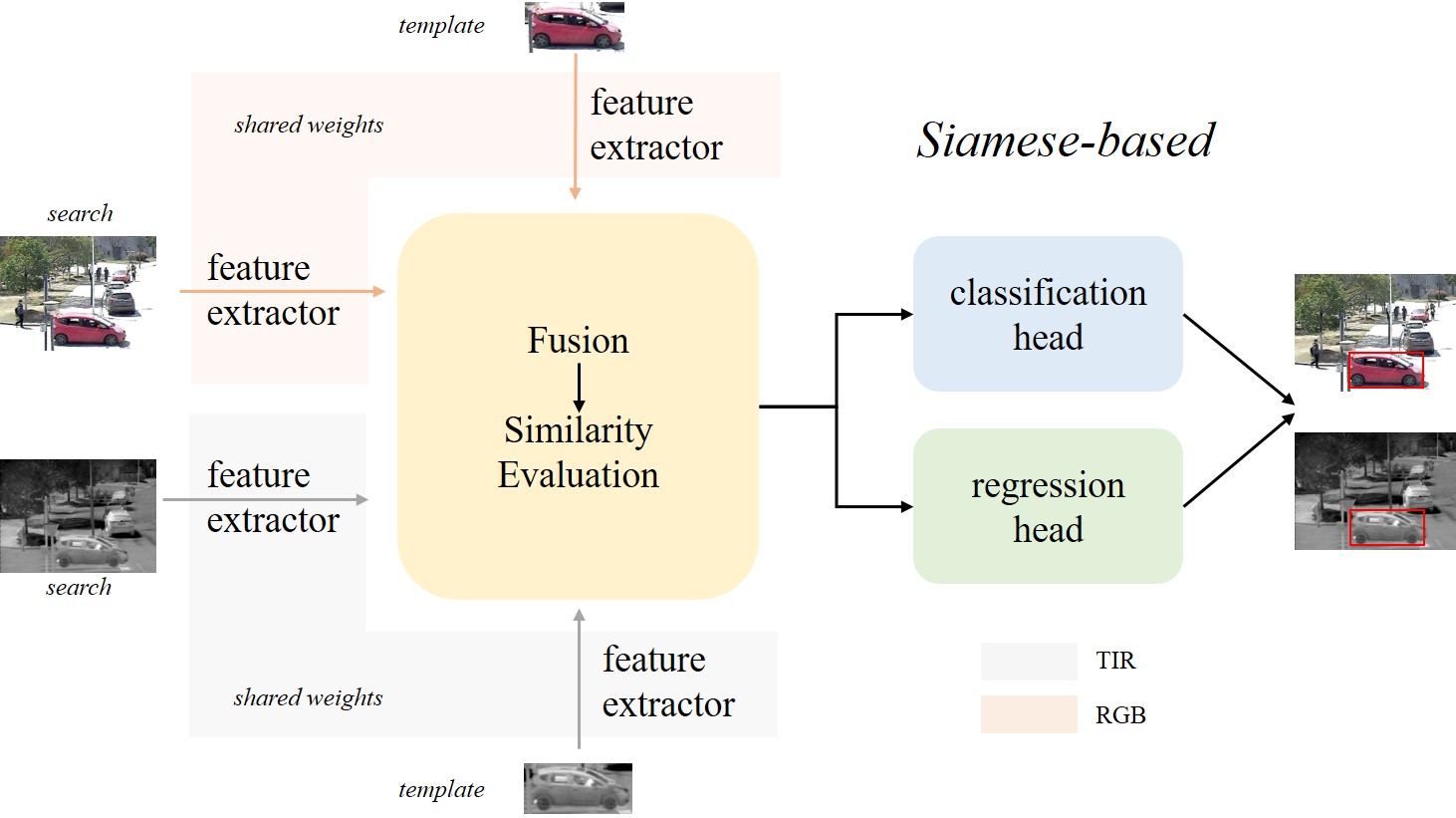} 
	\end{center}
    \caption{Pipeline of the published Siamese-based RGBT trackers. For fusion at feature level, the multi-modal fusion locates between the feature extraction and similarity evaluation. For pixel-level and decision-level fusion, the network architecture always degrades to a single-modality  configuration and, therefore, it is not included in this figure.}
    \label{fig:siamese-rgbt}
\end{figure*}

\subsection{MDNEt-based RGBT Trackers}\label{mdnet-based}
MDNet \cite{MDNet}, which is the champion of VOT-2015 challenge \cite{VOT2015}, treats each sequence as a single domain and uses the online updating technique to acquire the object-specific clues, obtaining robust tracking performance.
Thus, lots of RGBT trackers based on MDNet, which are presented in TABLE \ref{table:trackers}, is proposed.
And Fig \ref{fig:mdnet-rgbt} shows the basic extension of MDNet in RGBT tracking.
Generally, the feature extractor and fusion module are trained offline while the last fully-connected layer is trained online.
For example, at the beginning of the tracking procedure of 'Sequence 1', the first frame is utilized to train the parameters of the last fully-connected layer, giving more object-specific consideration.

Since the tracking process is generally similar, the main discrepancy of different algorithms lies in the feature extractor and fusion module, especially the latter.
Thus, we mainly focus on the fusion modules of the methods described below. 
In FANet \cite{FANet}, the feature channels from both RGB and TIR modalities are firstly concatenated for inter-modality interactions and then separated for fusion weights calculating, which is normalized by a Softmax operator applied in a cross-modal way.
DAPNet \cite{DAPNet} achieves the fusion task by recurrently using a sub-network, which consists of a convolutional layer, a Rectified Linear Unit (ReLU) activation \cite{ReLU} and a normalization layer, at different feature levels.
Compared with the coarse fusion sub-network used in DAPNet, DAFNet \cite{DAFNet} further designs an adaptive fusion module which is similar to that in FANet.
Carefully programmed, MANet \cite{MANet} expects to extract the modality-specific, modality-shared and object-specific clues through modality, generator and instance adapters respectively.
\cite{TODAT} investigates the potential of attention mechanism in RGBT tracking, \ie local attention for offline training and global attention for online testing.
Similarly, CMPP \cite{CMPP} employs the attention mechanism to finish the inter-modality correlation which is further extended to the temporal dimensional due to the significance of temporal information in video analysis.
A duality-gated mutual conditional module aims to achieve the multi-modal fusion task in a mutual-guided way in DMCNet \cite{DMCNet}.
Based on MANet, the features formerly enhanced within each modality before the fusion stage in MANet++ \cite{MANet++}.
A quality-aware fusion block is constructed to evaluate the modality significance in CBPNet \cite{CBPNet}.
Aiming at strengthen the stronger modality and suppress the weaker modality, a lightweight attention-based fusion module is applicated in M$^5$L \cite{M5L}.
All the above trackers mentioned in this sub-section maintain a fused feature representation after the cross-modal interactions.
However, to keep the modality-specific characteristics being discriminative, the features from RGB and TIR modalities are also retained in some methods.
Specifically, MacNet \cite{MaCNet} learns the fusion weights through the independent modal-aware attention network and competitive learning \cite{competitive-learning}, for which the features of RGB and TIR modalities are reserved, leverages the capacity of results from single modality and the modality-fused branch.
TFNet \cite{TFNet} deploys a trident branch architecture and each branch is specific for the RGB, TIR and fused features.
Different from the most existing RGBT trackers, CAT \cite{CAT}, ADRNet \cite{ADRNet} and APFNet \cite{APFNet} make their network construction more concrete for both modality-specific (\eg illumination variation in RGB and thermal crossover in TIR) and modality-shared (\eg scale variation) challenges.
In CAT \cite{CAT}, all the challenge-specific features are adaptively aggregated and then complemented to the basic learning procedure of both modalities.
ADRNet \cite{ADRNet} designs an attribute-driven residual block to measure the appearance model under different circumstance and then all the information is ensembled before the residual connection to the basic learned feature representation.
Similarly, in APFNet \cite{APFNet}, the aggregated multi-challenge features are combined with the features solely learned from RGB or TIR modality through transformer encoder and decoder blocks.

\subsection{Siamese-based RGBT Trackers}\label{siamese-based}
In visual object tracking, Siamese network is carried forward by the pioneering work SiamFC \cite{SiamFC} due to its efficiency brought by the end-to-end training scheme.
Totally, it aims to learn a general similarity evaluation metric and its pipeline is shown in Fig \ref{fig:siamese-rgbt}.
As the figure shows, the existing Siamese-based trackers employ unequal feature extractors for RGB and TIR modalities.
Then, in the published Siamese-based RGBT trackers, the multi-modal fusion module is followed to achieve the feature aggregation.
After that, the same strategy for similarity evaluation is applied for both classification and regression before their corresponding heads.

Since the core of multi-modal task lies in the combination of multi-modal clues, the fusion mechanism is mainly described for the following methods.
AT the beginning, the thermal data is employed by replacing one of the channels of RGB data in \cite{SiamRGT}.
Treating the SiamFC \cite{SiamFC} as the baseline, SiamFT \cite{SiamFT} uses simple concatenation for the features of template inputs while these of search inputs are fused by the learned modality reliabilities.
Based on SiamFT, the dynamic online-learned transformation strategy as well as the multi-level semantic features are further employed in DSiamMFT \cite{DSiamMFT}.
Similarly to \cite{SiamRGT}, fusion at the input stage, DuSiamRT \cite{DuSiamRT} utilizes modality-wise channel attention mechanism to fuse the features of template inputs while keeps unchanged for the features of search inputs.
SiamCDA \cite{SiamCDA}, whose baseline tracker is an advanced anchor-based tracekr, \ie SiamRPN++ \cite{SiamRPN++}, introduces the information fromm one modality to the other modality through the generated weights.
Furthermore, to cope with the situation at that time that there exists insufficient annotated RGBT data for large-scale network training, the LSS Dataset \footnote{https://github.com/RaymondCover/LSS-Dataset} is synthesized in a statistical way which contributes to its superior performance.

\subsection{Other Deep RGBT Trackers}
Except the MDNet-based and Siamese-based framework mentioned before, some RGBT trackers are built based on other frameworks.
In \cite{fusionnet}, the multi-modal information is straightforwardly combined by addition.
mfDiMP \cite{mfDimp} is the tracker based on DiMP \cite{Dimp} which is a superior RGB tracker that many researchers follow.
Specifically, DiMP is extended to the TIR modality and a TIR dataset is generated from GOT10k \cite{GOT-10K} and employed for the learning of neural network.
JMMAC \cite{JMMAC}, who is the champion of the published datasets of the VOT-RGBT2019 \cite{VOT2019} and VOT-RGBT2020 \cite{VOT2020}, learns the fusion weights through two sub-networks for local and global attention respectively.

\begin{table*}[ht]
\renewcommand\arraystretch{1.25}
\centering
\caption{\label{table:trackers}A collection of the existing deep RGBT trackers.}
\begin{tabular}{ccccc}
\toprule
\toprule
Trackers & Baseline & Year & Published                                 & Reference          \\               
\toprule
\toprule
   -     & Others         & 2018 & NeuroComputing                            & \cite{fusionnet} \\
 mfDiMP  & Others         & 2019 & ICCVW                                     & \cite{mfDimp}     \\
   MANet & MDNet-based    & 2019 & ICCVW                                     & \cite{MANet}    \\
  DAPNet & MDNet-based    & 2019 & ACM MM                                    & \cite{DAPNet}  \\
  DAFNet & MDNet-based    & 2019 & ICCVW                                     & \cite{DAFNet}  \\
   -     & MDNet-based    & 2019 & ICIP                                      & \cite{TODAT}  \\
   -     & Siamese-based  & 2019 & FUSION                                    & \cite{SiamRGT}  \\
 SiamFT  & Siamese-based  & 2019 & IEEE Access                               & \cite{SiamFT}  \\
 MaCNet  & MDNet-based    & 2020 & Sensors                                   & \cite{MaCNet} \\
 DMCNet  & MDNet-based    & 2020 & arXiv                                     & \cite{DMCNet}  \\
 CMPP    & MDNet-based    & 2020 & CVPR                                      & \cite{CMPP}   \\
 CAT     & MDNet-based    & 2020 & ECCV                                      & \cite{CAT}  \\
DSiamMFT & Siamese-based  & 2020 & Signal Processing: Image Communication    & \cite{DSiamMFT}  \\
JMMAC    & Others         & 2021 & IEEE TIP                                  & \cite{JMMAC}  \\
MANet++  & MDNet-based    & 2021 & IEEE TIP                                  & \cite{MANet++}  \\
CBPNet   & MDNet-based    & 2021 & IEEE TMM                                  & \cite{CBPNet}  \\
TFNet    & MDNet-based    & 2021 & IEEE TCSVT                                & \cite{TFNet}  \\
FANet    & MDNet-based    & 2021(2018) & IEEE TIV(arXiv)                     & \cite{FANet}  \\
ADRNet   & MDNet-based    & 2021 & IJCV                                      & \cite{ADRNet}  \\
M$^5$L   & MDNet-based    & 2021 & IEEE TIP                                  & \cite{M5L}  \\
SiamCDA  & Siamese-based  & 2021 & IEEE TCSVT                                & \cite{SiamCDA}  \\
DuSiamRT & Siamese-based  & 2021 & The Visual Computer                       & \cite{DuSiamRT}  \\
APFNet   & MDNet-based    & 2022 & AAAI                                      & \cite{APFNet}  \\

\bottomrule
\bottomrule
\end{tabular}
\end{table*}

\section{Datasets and Results}\label{datasetsandresults}
In this section, we will firstly give a introduction to the existing RGBT dataset, \ie GTOT \cite{GTOT}, RGBT210 \cite{RGBT210}, RGBT234 \cite{RGBT234}, VOT-RGBT2019 \cite{VOT2019}, VOT-RGBT2020 \cite{VOT2020} and LasHeR \cite{LasHeR}.
After that, a comparison of the existing deep RGBT trackers on multi-benchmarks will be listed and analysed.

\begin{table*}[ht]
\renewcommand\arraystretch{1.25}
\centering
\caption{\label{table:datasets}Illustration for RGBT datasets (Only for Testing).}
\begin{tabular}{ccccccc}
\toprule
\toprule
Benchmark   & Year   &  Num of Sequences & Aligned   & Category   & Num of attributes  & Reference \\
\toprule
\toprule
GTOT          &  2016   &  50  & N  & 9  & 7  & \cite{GTOT} \\
RGBT210       &  2017   & 210  & Y  & 22 & 12 & \cite{RGBT210} \\
RGBT234       &  2019   & 234  & N  & 22 & 12 & \cite{RGBT234} \\
VOT-RGBT2019  &  2019   & 60   & -  & 13 & 12 & \cite{VOT2019} \\
VOT-RGBT2020  &  2019   & 60   & -  & 13 & 12 & \cite{VOT2020} \\
LasHeR        & 2021    & 245  & Y  & 32 & 19 & \cite{LasHeR} \\
\bottomrule
\bottomrule
\end{tabular}
\end{table*}

\subsection{Datasets and Evaluation Metrics}\label{datasets}
TABLE \ref{table:datasets} shows the detail information about these six benchmarks.
Here 'Num of Sequences' represents the number of paired RGBT videos.
'Aligned' means whether the RGB and TIR images are aligned or not.
Specifically, if there exists one groundtruth file ('Y'), the RGB and TIR modalities are thought aligned. 
It should be noticed that the VOT-RGBT2019 dataset is a sub-set of RGBT234 and the difference between VOT-RGBT2019 and VOT-RGBT2020 benchmarks locate in the testing protocol.
Therefore, they are the same in statistic analysis \cite{VOT2020}.

The same evaluation metrics are employed in GTOT \cite{GTOT}, RGBT210 \cite{RGBT210}, RGBT234 \cite{RGBT234} and LasHeR \cite{LasHeR} datasets, \ie Precision and Success.
Precision rate measures the distance between the groundtruth bounding box and the predicted one.
Success rate represents the ratio of tracking failures whose Intersection over Union (IoU) between its corresponding label below a given threshold.

Accuracy, Robustness and (Excepted Average Overlap) EAO are the measurements utilized in VOT-RGBT2019 \cite{VOT2019} and VOT-RGBT2020 \cite{VOT2020}.
The overlap between the prediction and the groundtruth is represented by accuracy.
Robustness is designed to measure the ratio of tracking failures over the total numbers of image frames.
EAO is considered the most important and comprehensively indicates the superiority of the tracker.

\begin{table*}[ht]
\renewcommand\arraystretch{1.25}
\centering
\caption{\label{table:results}Quantitative results of the existing deep RGBT trackers on GTOT, RGBT210, RGBT234 and LasHeR datasets.}
\begin{tabular}{ccccccccc}
\toprule
\toprule
         & \multicolumn{2}{c}{GTOT \cite{GTOT}} & \multicolumn{2}{c}{RGBT210 \cite{RGBT210}} & \multicolumn{2}{c}{RGBT234 \cite{RGBT234}} & \multicolumn{2}{c}{LasHeR \cite{LasHeR}} \\ 
Trackers & Precision$(\uparrow)$ & Success$(\uparrow)$ & Precision$(\uparrow)$ & Success$(\uparrow)$ & Precision$(\uparrow)$ & Success$(\uparrow)$ & Precision$(\uparrow)$ & Success$(\uparrow)$ \\
\toprule
\toprule
   \cite{fusionnet}     & 0.852  & 0.626 & -      & -      & -      & -      & -      & -     \\
 mfDiMP \cite{mfDimp}   & _      & -     & 0.786  & 0.555  & 0.785  & 0.559  & 0.447  & 0.344 \\
   MANet \cite{MANet}   & 0.894  & 0.724 & -      & -      & 0.777  & 0.539  & 0.457  & 0.33  \\ 
  DAPNet \cite{DAPNet}  & 0.882  & 0.707 & -      & -      & 0.766  & 0.537  & 0.431  & 0.314 \\
  DAFNet \cite{DAFNet}  & 0.891  & 0.712 & -      & -      & 0.796  & 0.544  & 0.449  & 0.311 \\
   \cite{TODAT}         & 0.843  & 0.677 & -      & -      & 0.787  & 0.545  & -      & -     \\
   \cite{SiamRGT}       & -      & -     & -      & -      & 0.610  & 0.428  & -      & -     \\
 SiamFT \cite{SiamFT}   & 0.826  & 0.700 & -      & -      & 0.688  & 0.486  & -      & -     \\
 MaCNet \cite{MaCNet}   & 0.880  & 0.714 & -      & -      & 0.790  & 0.554  & 0.483  & 0.352 \\
 DMCNet  \cite{DMCNet}  & 0.909  & 0.733 & 0.797  & 0.555  & 0.839  & 0.593  & 0.491  & 0.357 \\
 CMPP   \cite{CMPP}     & 0.926  & 0.738 & -      & -      & 0.823  & 0.575  & -      & -     \\
 CAT    \cite{CAT}      & 0.889  & 0.717 & 0.792  & 0.533  & 0.804  & 0.561  & 0.451  & 0.317 \\
DSiamMFT \cite{DSiamMFT}& -      & -     & 0.642  & 0.432  & -      & -      & -      & -     \\
JMMAC    \cite{JMMAC}   & 0.902  & 0.732 & -      & -      & 0.790  & 0.573  & -      & -     \\
MANet++  \cite{MANet++} & 0.901  & 0.723 & -      & -      & 0.800  & 0.554  & 0.467  & 0.317 \\
CBPNet   \cite{CBPNet}  & 0.885  & 0.716 & -      & -      & 0.794  & 0.541  & -      & -     \\
TFNet    \cite{TFNet}   & 0.886  & 0.729 & 0.777  & 0.529  & 0.806  & 0.560  & -      & -     \\
FANet    \cite{FANet}   & 0.891  & 0.728 & -      & -      & 0.787  & 0.553  & 0.442  & 0.309 \\
ADRNet   \cite{ADRNet}  & 0.904  & 0.739 & -      & -      & 0.809  & 0.571  & -      & -     \\
M$^5$L   \cite{M5L}     & 0.896  & 0.710 & -      & -      & 0.795  & 0.542  & -      & -     \\
SiamCDA  \cite{SiamCDA} & 0.877  & 0.732 & -      & -      & 0.760  & 0.569  & -      & -     \\
DuSiamRT \cite{DuSiamRT}& 0.766  & 0.628 & -      & -      & 0.567  & 0.384  & -      & -     \\
APFNet   \cite{APFNet}  & 0.905  & 0.739 & -      & -      & 0.827  & 0.579  & 0.500  & 0.362 \\
\bottomrule
\bottomrule
\end{tabular}
\end{table*}

\begin{table*}[ht]
\renewcommand\arraystretch{1.25}
\centering
\caption{\label{table:results-vot}Quantitative results on VOT-RGBT2019 and VOT-RGBT2020 datasets.}
\begin{tabular}{ccccccc}
\toprule
\toprule
         & \multicolumn{3}{c}{VOT-RGBT2019 \cite{VOT2019}} & \multicolumn{3}{c}{VOT-RGBT2020 \cite{VOT2020}} \\ 
Trackers & Accuracy$(\uparrow)$ & Robustness$(\uparrow)$ & EAO$(\uparrow)$ & Accuracy$(\uparrow)$ & Robustness$(\uparrow)$ & EAO$(\uparrow)$ \\
\toprule
\toprule
 mfDiMP \cite{mfDimp}   & 0.6019 & 0.8036 & 0.3879 & 0.6380 & 0.7930 & 0.3800  \\
   MANet \cite{MANet}   & 0.5823 & 0.7010 & 0.3463 & -      & -      & -       \\
 SiamFT \cite{SiamFT}   & 0.6300 & 0.6390 & 0.3100 & -      & -      & -       \\
 MaCNet \cite{MaCNet}   & 0.5451 & 0.5914 & 0.3052 & -      & -      & -       \\
JMMAC    \cite{JMMAC}   & 0.6649 & 0.8211 & 0.4826 & 0.6620 & 0.8180 & 0.4200  \\ 
MANet++  \cite{MANet++} & 0.5092 & 0.5379 & 0.2716 & -      & -      & -       \\
TFNet    \cite{TFNet}   & 0.4617 & 0.5936 & 0.2878 & -      & -      & -       \\
FANet    \cite{FANet}   & 0.4724 & 0.5078 & 0.2465 & -      & -      & -       \\
ADRNet   \cite{ADRNet}  & 0.6218 & 0.7657 & 0.3959 & -      & -      & -       \\
SiamCDA  \cite{SiamCDA} & 0.6820 & 0.7570 & 0.4240 & -      & -      & -       \\
\bottomrule
\bottomrule
\end{tabular}
\end{table*}

\subsection{Results}\label{results}
TABLE \ref{table:results} shows the results on GTOT \cite{GTOT}, RGBT210 \cite{RGBT210}, RGBT234 \cite{RGBT234} and LasHeR \cite{LasHeR} while the results on the VOT banechmarks are exhibited in TABLE \ref{table:results-vot}.
On GTOT, the highest Success rate is obtained by ADRNet \cite{ADRNet} and AFPNet \cite{APFNet} (0.739) while the best Precision rate reaches 0.926 by CMPP \cite{CMPP}.
On RGBT210, DMCNet \cite{DMCNet} achieves the best Precision (0.797) and Success (0.555) rates at the same time.
Consistently, DMCNet also ranks the first on Precision (0.839) and Success (0.593) rates on RGBT234 dataset.
APFNet \cite{APFNet} gets the best Precision and Success scores on LasHeR dataset.
For the VOT benchamrks, as mentioned before, JMMAC \cite{JMMAC} ranks the first on the public dataset twice.
However, the VOT community provides one public dataset combined with a sequestered one, and the real champion is decided on the private dataset.
The champion of VOT-RGBT2019 \cite{VOT2019} is mfDiMP \cite{mfDimp} and DFAT wins the VOT-RGBT2020 challenge \cite{VOT2020}.

\section{Discussion}\label{discussion}
From the above investigations, we have several discussions as follows:
(1) Following the whole computer vision field, the potential of Transformer model \cite{transformer} is not explored yet.
(2) Different the image-based tasks \cite{li2011no, chen2018new, zheng2006nearest, li2020nestfuse, feng2017face, wu2004new, sun2019effective, wang2003initial, sun2011quantum}, tempoal information , which has not been widely studied in RGBT tracking yet, is of great inportance in video-based tasks, \eg visual object tracking.
(3) During the investigation, we find that less mathematical theories considered during the network construction process.
(4) Behind the fusion step, what is actually going on has not been discussed yet.
In the future, we will mainly focus on more concrete RGBT tracking based on our discussions.
\section{Conclusion}\label{conclusion}
In this paper, a statistic analysis of the existing deep RGBT trackers.
Specifically, all the trackers are divided into three categories, \ie MDNet-based, Siamese-based and Others.
Furthermore, their quantitative results on GTOT, RGBT210, RGBT234, LasHeR, VOT-RGBT2019 and VOT-RGBT2020 benchmarks are compared intuitively by gathering them together.
Therefore, this work can act like a reference for researchers who are interested in RGBT tracking.




\bibliographystyle{IEEEtran} 
\bibliography{sample} 
\end{document}